\documentclass[10pt,twocolumn,letterpaper]{article}

\usepackage{cvis}              

\usepackage[dvipsnames]{xcolor}
\usepackage{graphicx}

\usepackage{array}
\usepackage{amssymb}
\usepackage{amsmath}
\usepackage{pifont}
\newcommand{\xmarkgrey}{{\color{gray}\ding{55}}}

\graphicspath{{figures/}{attn_maps/}}

\definecolor{cvisblue}{rgb}{0.21,0.49,0.74}
\usepackage[pagebackref,breaklinks,colorlinks,allcolors=cvisblue]{hyperref}
\def\confName{CVIS}
\def\confYear{2026}

\title{Understanding vision transformer quantization robustness through the lens of out-of-distribution detection}

\author{Joey Kuang, Alexander Wong\\
Vision and Image Processing Research Group, Department of Systems Design Engineering\\
University of Waterloo, Ontario, Canada\\
{\tt\small \{jjykuang,a28wong\}@uwaterloo.ca}
}

\begin{document}
\maketitle
\begin{abstract}
Vision transformers have shown remarkable performance in vision tasks, but enabling them for accessible and real-time use is still challenging. Quantization reduces memory and inference costs at the risk of performance loss. Strides have been made to mitigate low precision issues mainly by understanding in-distribution (ID) task behaviour, but the attention mechanism may provide insight on quantization attributes by exploring out-of-distribution (OOD) situations. We investigate the behaviour of quantized small-variant popular vision transformers (DeiT, DeiT3, and ViT) on common OOD datasets. ID analyses show the initial instabilities of 4-bit models, particularly of those trained on the larger ImageNet-22k, as the strongest FP32 model, DeiT3, sharply drops 17\% from quantization error to be one of the weakest 4-bit models. While ViT shows reasonable quantization robustness for ID calibration, OOD detection reveals more: ViT and DeiT3 pretrained on ImageNet-22k respectively experienced a 15.0\% and 19.2\% average quantization delta in AUPR-out between full precision to 4-bit while their ImageNet-1k-only counterparts experienced a 9.5\% and 12.0\% delta. Overall, our results suggest pretraining on large scale datasets may hinder low-bit quantization robustness in OOD detection and that data augmentation may be a more beneficial option.

\end{abstract}    
\section{Introduction}
\label{sec:intro}

In recent years, vision transformers (ViTs) have risen to be top performers in large-scale vision tasks and are becoming more widely considered as a new standard in deep computer vision. They rely on the necessarily quadratic self-attention mechanism \cite{pmlr-v201-duman-keles23a}, employing multiple heads of attention in one block, with several blocks in its large backbone. This makes ViTs poor candidates for edge deployment. However, model compression techniques such as quantization may enable ViTs for more powerful real-time use cases.

\begin{figure}[t]
   \centering
   \begin{subfigure}{\columnwidth}
    \centering
    \includegraphics[width=0.8\textwidth]{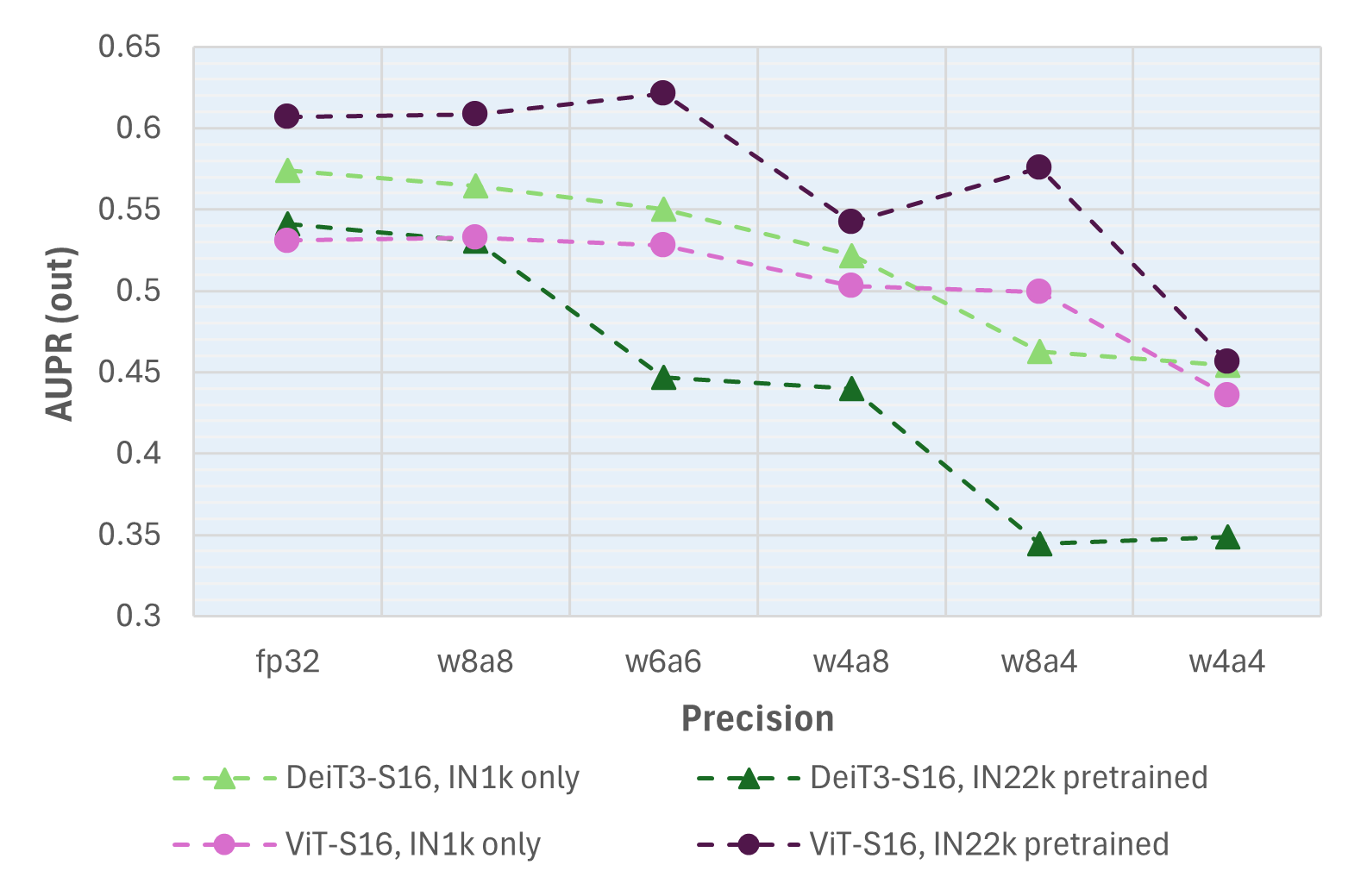}
    \caption{AUPR-out averaged across five OOD datasets (all listed in Sec. \ref{sec:methodology}, except for ImageNet-O. Average random chance level is 0.175.}
   \end{subfigure}
   \begin{subfigure}{\columnwidth}
    \centering
    \includegraphics[width=0.8\textwidth]{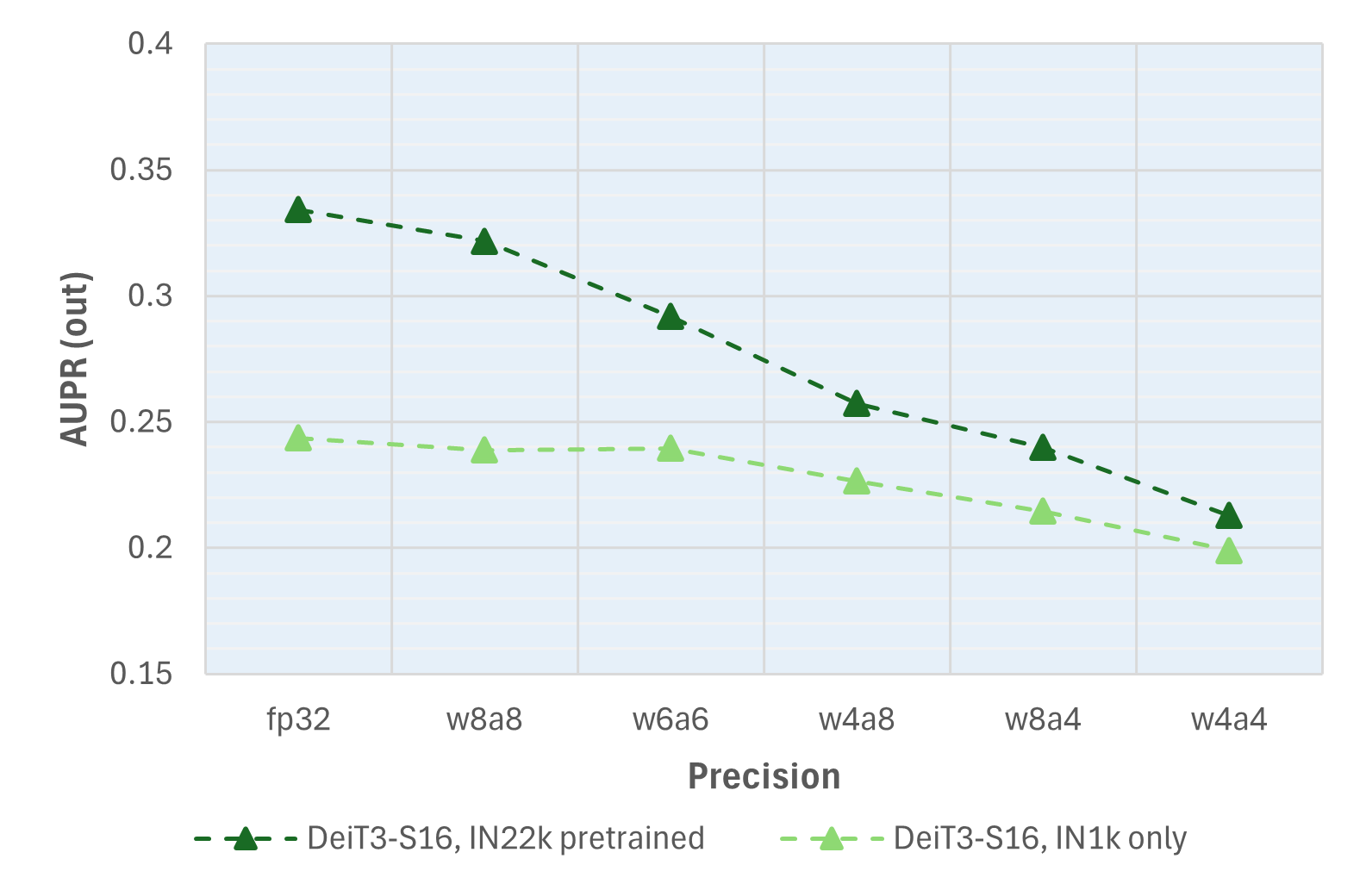}
    \caption{AUPR-out on ImageNet-O. Random chance level is 0.167.}
   \end{subfigure}
   \caption{Higher performing models assign OOD examples with higher anomaly scores. See Sec. \ref{sec:methodology} for a description of the AUPR-out as an OOD detection measure. Models pretrained on the larger dataset are more sensitive to quantization error in OOD detection.}
   \label{fig:aupr_out_comparisons}
\end{figure}


\begin{table*}[t]
    \centering
    \resizebox{\textwidth}{!}{%
    \begin{tabular}{lcccccccccccc}
    \toprule
        \multicolumn{1}{l|}{} &
                                         \multicolumn{6}{c|}{Top-1 accuracy (\%)}             & \multicolumn{6}{c}{NLL}                       \\ \midrule
    \multicolumn{1}{l|}{} &
      \multicolumn{1}{c}{FP32} &
      \multicolumn{1}{c}{W8A8} &
      \multicolumn{1}{c}{W6A6} &
      \multicolumn{1}{c}{W4A8} &
      \multicolumn{1}{c}{W8A4} &
      \multicolumn{1}{c|}{W4A4} &
      \multicolumn{1}{c}{FP32} &
      \multicolumn{1}{c}{W8A8} &
      \multicolumn{1}{c}{W6A6} &
      \multicolumn{1}{c}{W4A8} &
      \multicolumn{1}{c}{W8A4} &
      \multicolumn{1}{c}{W4A4} \\ \midrule
    \multicolumn{1}{l|}{\textbf{DeiT3-S16, IN22k pretrained}} & 82.700 & 82.594 & 81.380 & 73.491 & 76.287 & \multicolumn{1}{c|}{65.288} & 0.733 & 0.819 & 1.002 & 1.171 & 1.325 & 1.662 \\
    \multicolumn{1}{l|}{\textbf{DeiT3-S16, IN1k only}}        & 81.302 & 81.212 & 80.144 & 74.768 & 74.052 & \multicolumn{1}{c|}{66.873} & 0.894 & 0.897 & 0.974 & 1.334 & 1.440 & 1.927 \\
    \multicolumn{1}{l|}{\textbf{ViT-S16, IN22k pretrained}}   & 81.337 & 81.151 & 80.265 & 75.484 & 73.173 & \multicolumn{1}{c|}{64.812} & 0.673 & 0.682 & 0.726 & 0.926 & 1.057 & 1.486 \\
    \multicolumn{1}{l|}{\textbf{DeiT-S16, IN1k only}}         & 79.786 & 79.714 & 78.829 & 74.831 & 73.367 & \multicolumn{1}{c|}{67.933} & 0.883 & 0.888 & 0.945 & 1.148 & 1.242 & 1.512 \\ \bottomrule
    \end{tabular}%
    }
    \caption{Quantization performance on ImageNet-1k (in-distribution). Each datapoint is an average across three seeds. All models have the same architecture and only differentiate by training method and dataset, but demonstrate varying levels of robustness to lower precision.}
    \label{tab:top1_nll_markers}

\end{table*}

\begin{table*}[t]
    \centering
    \resizebox{\textwidth}{!}{%
    \begin{tabular}{ccccc}
        \toprule
        \multicolumn{1}{l|}{} &
          VIT-S16, IN22k \cite{steiner_how_2022} &
          DeiT-S16, IN1k \cite{touvron_training_2021} &
          DeiT3-S16, IN1k \cite{touvron_deit_2022} &
          Deit3-S16, IN22k \cite{touvron_deit_2022} \\ \midrule
        \multicolumn{1}{l|}{Batch size}                 & 4096                                                                              & 1024       & 2048       & 2048       \\
        \multicolumn{1}{l|}{Num epochs (PT/FT)}         & 300/30  & 300        & 400        & 90/50      \\
        \multicolumn{1}{l|}{Optimizer}                  & AdamW                                                                             & AdamW      & LAMB       & LAMB       \\
        \multicolumn{1}{l|}{Weight decay}               & 0.1                                                                               & 0.05       & 0.02       & 0.02       \\ \midrule
        \multicolumn{1}{l|}{Label smoothing $\epsilon$} & 0.1                                                                               & 0.1        & \xmarkgrey & 0.1        \\
        \multicolumn{1}{l|}{Gradient Clip}              & Norm, 1.0                                                                         & \xmarkgrey & Norm, 1.0  & Norm, 1.0  \\ \midrule
        \multicolumn{1}{l|}{RRC}                        & \checkmark                                                                        & \checkmark & \checkmark & \xmarkgrey \\
        \multicolumn{1}{l|}{RandAugment ($N$/$M$/$M_{std}$)}                & 2/15/0.0                                                                          & 2/9/0.5    & \xmarkgrey & \xmarkgrey \\
        \multicolumn{1}{l|}{3 Augment}                  & \xmarkgrey                                                                        & \xmarkgrey & \checkmark & \checkmark \\
        \multicolumn{1}{l|}{LayerScale}                 & \xmarkgrey                                                                        & \xmarkgrey & 1e-4       & 1e-4       \\
        \multicolumn{1}{l|}{Mixup alpha}                & 0.2                                                                               & 0.8        & 0.8        & \xmarkgrey \\
        \multicolumn{1}{l|}{Cutmix alpha}               & \xmarkgrey                                                                        & 1.0        & 1.0        & 1.0        \\
        \multicolumn{1}{l|}{Erasing prob.}              & \xmarkgrey                                                                        & 0.25       & \xmarkgrey & \xmarkgrey \\
        \multicolumn{1}{l|}{ColorJitter}                & \xmarkgrey                                                                        & \xmarkgrey & 0.3        & 0.3        \\ \bottomrule
    \end{tabular}%
    }
    \caption{Summary of training procedures for the models evaluated. For ViT by \cite{steiner_how_2022}, note that the model trained on ImageNet-1k only is also trained for 300 epochs and the configuration was taken from the specific model checkpoint\protect\footnotemark. We show this table to highlight differences in training methods that may contribute to our observations in quantized OOD performance.}
    \label{tab:summary_of_training_procedures}
\end{table*}

Quantization compresses a model by reducing the precision required to represent the parameters, reducing both memory overhead and computational complexity. Lowering the precision introduces noise that may cause performance deltas. There are two general frameworks of quantization: quantization-aware training (QAT) and post-training quantization (PTQ). QAT involves fully re-training the model with objectives that minimize the quantized model error. This is often resource-consuming and challenging to configure \cite{jacob_quantization_2017}. PTQ is a less precise, but more practical solution, requiring only a small calibration dataset to determine quantization parameters \cite{nagel_white_2021}. While convolutional neural networks (CNNs) have enjoyed success from techniques developed in this field, ViTs could not see translated benefits. Early works exploring ViT quantization have observed varying complex distributions in LayerNorm, Softmax, and GeLU activations (the former two being part of the attention block), requiring the proposal of new PTQ schemes \cite{lin_fq-vit_2022,li_repq-vit_2023,avidan_ptq4vit_2022,ding_towards_2022}. 

While there is commendable progress in closing the quantization gap for accuracy, we also wonder if the same can be said for out-of-distribution (OOD) tasks. If we consider ViTs for deployment, then real-time usage will challenge models with unseen, noisy inputs. In their full precision form, the features produced by attention blocks have shown to be valuable in general robustness to distribution shifts, citing lack of inductive bias allowing for richer representation learning \cite{fort_exploring_2021,cultrera_leveraging_2023,sim_attention_2023,minderer_calibration_2021}. However, Pinto \etal \cite{pinto_are_2021} discuss the role of pretraining actually being more significant than the inductive bias when comparing ViTs to CNNs. In any case, quantization integrity in the attention block is ostensibly crucial to its performance for any distribution. 


We investigate whether OOD-related tasks can add insight on common vision transformer robustness to quantization. We test quantized models on six common OOD datasets and evaluate their ability to distinguish OOD samples from in-distribution (ID) ones. Results shown in Fig. \ref{fig:aupr_out_comparisons} suggests that large-scale pretraining hinders low-bit quantization robustness and this is made apparent in OOD detection. This leads us to visualize attention to see whether our quantitative results align with our expectations in quantized attention integrity. Our findings suggest that data augmentation is beneficial for OOD tasks even in lower precision.


\footnotetext{\path{https://storage.googleapis.com/vit_models/augreg/[S_16-i21k-300ep-lr_0.001-aug_medium2-wd_0.1-do_0.0-sd_0.0--imagenet2012-steps_20k-lr_0.03-res_224.npz, S_16-i1k-300ep-lr_0.001-aug_medium2-wd_0.1-do_0.0-sd_0.0.npz]}}

\section{Methodology}
\label{sec:methodology}

\subsection{Motivation}

The vision transformer architecture embeds image patches which are fed into blocks, each composed of a multi-head self-attention (MHSA) module and a multi-layer perceptron module. The MHSA projects these tokenized patches into three representations (query, key, value) and applies Softmax to represent attention scores in $[0, 1]$. These attention scores aim to inform the model of relevant inter-patch correlations; preserving attention features in quantization is therefore crucial in vision tasks. We propose that this goes beyond maintaining the performance in classifying ID samples, but also preserving learned features that help the model distinguish OOD samples from ID. 

We would like to probe the quantized vision transformer's behaviour in both an ID and OOD setting. If the behaviour is aligned between distribution shifts, then the current paradigm of benchmarking quantization performance can likely continue as is. Otherwise, we may gain new insight on how we can approach vision transformer quantization. To narrow down the model behaviour in OOD contexts, we specifically look at models that differ only by training protocol; previous works have discussed the value of certain training techniques in model robustness \cite{hendrycks_using_2019, geiping_how_2022}. 

We start with a brief analysis on an ID task using classification accuracy and negative-log likelihood (NLL), common metrics that practictioners use to benchmark quantization robustness. The latter is a proper score and strong indicator of model calibration, which is relevant for robustness to different distributions. Since we are hypothesizing that quantization of the attention mechanism can lead to a deterioration in learned correlations, we should ask how well can the model distinguish a sample as OOD when it correctly identifies it as so vs. when it incorrectly detects as so. We use maximum softmax probability, suggested by Hendrycks \etal \cite{hendrycks_baseline_2022}, to obtain the Area Under Precision-Recall (AUPR) scores for OOD samples (OOD as the 'positive' class and ID as the 'negative'). The following section describes the experiment setup in detail.  





\subsection{Setup}
We investigate small-variant vision transformers of the same architecture, either pretrained on ImageNet-22k (IN22k) \cite{deng_imagenet_2009} or ImageNet-1k (IN1k) \cite{russakovsky_imagenet_2015}: ViT \cite{steiner_how_2022}, DeiT \cite{touvron_training_2021}, and DeiT3 \cite{touvron_deit_2022}. These size variants are commonly studied in other quantization works, and are more relevant in the context of edge devices (albeit still massive for edge). Model weights are obtained from Timm library\footnote{https://github.com/huggingface/pytorch-image-models/tree/main/timm} and from Google Research's cloud storage.


We quantize all networks using the open-sourced RepQ-ViT \cite{li_repq-vit_2023,li_repquant_2024}, a strong PTQ method for ViTs. We follow the original paper's quantization configuration. Seeds 0, 13, 37 are used to generate a robust set of results.
For evaluation on ID performance, we use ImageNet-1k. For OOD evaluation, we use six commonly tested datasets: ImageNet-O \cite{hendrycks_natural_2021}, Textures \cite{cimpoi_describing_2014}, Places365 \cite{zhou_places_2018}, SUN \cite{xiao_sun_2010}, iNaturalist \cite{van_horn_inaturalist_2018}, and OpenImage-O \cite{wang_vim_2022}. 
The baseline ID analysis reports Top-1 accuracy and NLL for insight into quantization errors related to task performance and calibration. We report AUPR-out \cite{hendrycks_baseline_2022} scores to evaluate OOD robustness, using ImageNet-1k validation set as the ID comparison set. For attention visualization, we use attention rollout and follow Gildenblat's suggestion on discarding lowest pixels (ratio of 0.9) and taking the maximum value among the heads for best visualization \cite{abnar_quantifying_2020,noauthor_exploring_2020}.

\section{Key findings}
\label{sec:keyfindings}

\newlength{\imgwidth}
\setlength{\imgwidth}{2.0cm}
\renewcommand{\arraystretch}{1.2}

\begin{figure*}[t]
    \centering
    \begin{subfigure}{0.47\textwidth}
    \footnotesize
        \centering
        {
            \setlength{\tabcolsep}{2pt}
            \begin{tabular}{m{3.5em}>{\centering\arraybackslash}m{\imgwidth}>{\centering\arraybackslash}m{\imgwidth}>{\centering\arraybackslash}m{\imgwidth}}
                Prec. $\rightarrow$ & FP32 & W4A8 & W4A4 \\ 
                DeiT, IN1k & \includegraphics[width=\imgwidth]{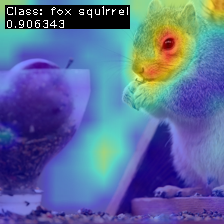} & 
                \includegraphics[width=\imgwidth]{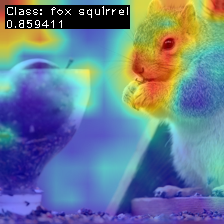} & 
                \includegraphics[width=\imgwidth]{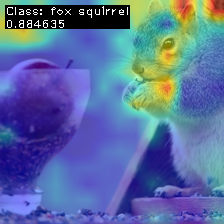} \\ 
                DeiT3, IN1k & \includegraphics[width=\imgwidth]{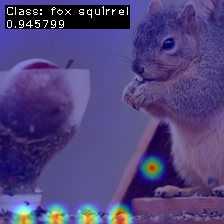} &
                \includegraphics[width=\imgwidth]{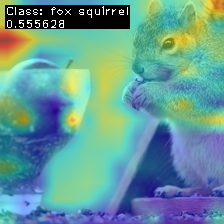} & 
                \includegraphics[width=\imgwidth]{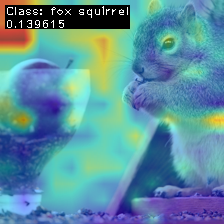} \\ 
                DeiT3, IN22k & \includegraphics[width=\imgwidth]{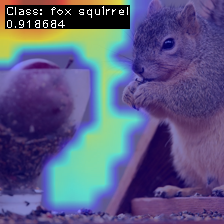} & 
                \includegraphics[width=\imgwidth]{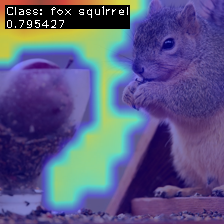} & 
                \includegraphics[width=\imgwidth]{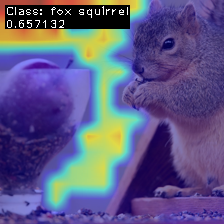}
            \end{tabular}
        }
        \caption{Attention rollout maps for an ID sample (ImageNet-1k).}
    \end{subfigure}
    \begin{subfigure}{0.45\textwidth}
    \footnotesize
        \centering
        {
            \setlength{\tabcolsep}{2pt}
            \begin{tabular}{>{\centering\arraybackslash}m{\imgwidth}>{\centering\arraybackslash}m{\imgwidth}>{\centering\arraybackslash}m{\imgwidth}} 
            FP32\vphantom{DeiT3, IN22k} & W4A8 & W4A4 \\
            \includegraphics[width=\imgwidth]{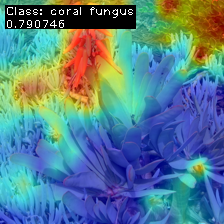} & 
            \includegraphics[width=\imgwidth]{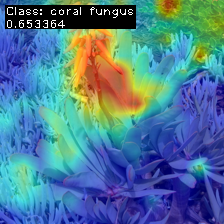} & 
            \includegraphics[width=\imgwidth]{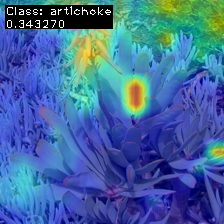} \\
            \includegraphics[width=\imgwidth]{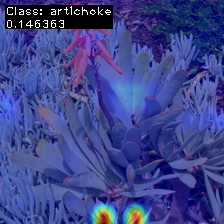} &
            \includegraphics[width=\imgwidth]{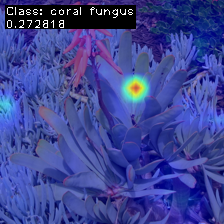} & 
            \includegraphics[width=\imgwidth]{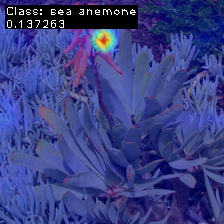} \\
            \includegraphics[width=\imgwidth]{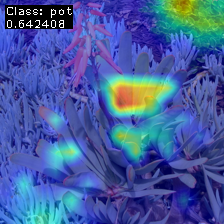} & 
            \includegraphics[width=\imgwidth]{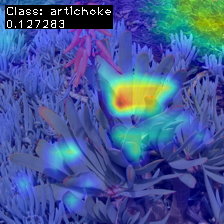} & 
            \includegraphics[width=\imgwidth]{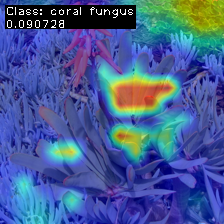}
            \end{tabular}
        }
        \caption{Attention rollout maps for an OOD sample (iNaturalist).}
    \end{subfigure}
    \caption{DeiT \cite{touvron_training_2021} produces interpretable attention maps even at lower precision. We observe outlier artifacts similar to findings in \cite{darcet_vision_2023} in DeiT3 features. Despite poor AUPR robustness to 4-bit quantization, attention maps produced by DeiT3-S pretrained on IN22k seem stable. Attention maps do not appear to vary greatly under OOD data. Predicted class label and probability are labeled in each image.}
    \label{fig:attn_maps}
\end{figure*}

\subsection{Vision transformers have unpredictable levels of robustness to 4-bit quantization}

ViT-likes are fairly robust to 8- and 6-bit quantization under the RepQ-ViT framework. Once the model faces 4-bit quantization, we start to see instability in the highest performing models. Tab. \ref{tab:top1_nll_markers} shows the strongest FP32 model, DeiT3-S16 (IN22k pretrained), steeply drop over 17\% in accuracy at full 4-bit. ViT-S16 (IN22k pretrained) suffers a 16.5\% drop, while the two IN1k-trained models are better able to retain classification performance at 4-bits (DeiT3 at 14.4\% and DeiT at 11.9\%).
The differences in pretraining scale do not explain every result, as both DeiT3 models suffer a sharp increase in NLL. Meanwhile, ViT-S16 is consistently the most calibrated model at every precision, and DeiT-S shows minimal quantization error again.

Since all models have the same architecture, we can hone in on training procedure. We highlight the training methods between the four models in Tab. \ref{tab:summary_of_training_procedures}. Of note for DeiT is the lack of gradient clipping and the large amount of data augmentation applied. Gradient clipping stabilizes the optimization process in training, nearly guaranteeing convergence \cite{pascanu_difficulty_2013,mai_stability_2021}. If this technique lends to poor 4-bit stability, it implies that optimal model parameters in their trained data range (FP32) have severely high variance. More plausible is the idea that DeiT's aggressive use of data augmentation have flattened their loss landscape, an observation explored by \cite{geiping_how_2022}, which seems to have translated in quantization. The remainder of the investigation focuses on the two pairs of IN1k-trained and IN22k-pretrained models by \cite{steiner_how_2022} and \cite{touvron_deit_2022} to further scrutinize effects of pretraining scale and other training decisions. 


\subsection{Vision transformers pretrained on ImageNet-22k are less robust to quantization in OOD}

We now analyze ViT and DeiT3 in OOD scenarios. Although pretraining was not a performance indicator for the ID case, we see a more consistent pattern in quantization stability for OOD tests. The IN22k-pretrained models sharply drop in their ability to distinguish OOD samples from ID, regardless of their capabilities at FP32.
In Fig. \ref{fig:aupr_out_comparisons}(a), the 3.3\% gap between the two FP32-DeiT3 models widens to 10.5\% at 4-bit, and the IN1k variant consistently outperforms the IN22k-pretrained variant. We find a similar behaviour in stability on ImageNet-O. This is significant since the IN22k-pretrained model has actually seen this data (ImageNet-O is constructed from a subset of IN22k), so we expect a better ability to distinguish it as OOD. Tab. \ref{tab:summary_of_training_procedures} shows the additional augmentations of RRC and Mixup on the side of the IN1k-trained model. This further supports the benefits of data augmentations for loss landscapes translating under quantized settings. 


We also see that a 7.6\% AUPR-out gap between the two FP32-ViT models diminishes to 2.1\% at 4-bit. In this case, the only difference in training is the data seen and total time training, again suggesting that large-scale pretraining may have adverse effects to OOD data in the low-bit regime. 


\subsection{High-norm outlier tokens in attention maps may be too dominant in lower precision}
To illustrate the effects of quantization on vision transformers, we visualize the attention attribution at each precision for the DeiT and DeiT3 models in Fig. \ref{fig:attn_maps}. DeiT3 features contain a lot of artifact patches (hot spots in irrelevant areas) resembling the phenomenon of high-norm outlier tokens discussed by \cite{darcet_vision_2023}. Darcet \etal \cite{darcet_vision_2023} investigated this phenomenon and suggested that pretraining at a large scale for longer durations and larger model sizes can lead to forming outlier tokens. They also find that these outlier tokens contain global information. Given that we use the Small variant, the training duration of DeiT3 is a plausible contributor in learning these tokens.


Interestingly, DeiT3-S-IN22k attention maps appear to be robust at 4-bits; in the IN1k-trained counterpart maps, the outlier tokens "move" across the image, suggesting sensitivity to 4-bits. While this aligns with the quantization error in ID calibration, the changes in outlier attributes did not hinder OOD detection capabilities. Yellapragada \etal \cite{yellapragada_leveraging_2025} show that the learned representations of high-norm tokens distilled into registers do not generalize well in OOD. The IN22k-pretrained model appears to have retained the global information of these outlier tokens at lower bits, but it is possible that other local representations may have been saturated. Meanwhile, the outlier norm in the IN1k model may not have been severe, which causes quantization error in the global information at lower bits (adapting the range to a lower magnitude), but other information useful for anomaly detection is better preserved.
\section{Conclusion}
\label{sec:conclusion}

Efforts to improve vision transformer performance in the quantized regime have not necessarily translated in OOD scenarios. We present findings that suggest that pretraining and data augmentation play a non-trivial role in dictating model stability in low-bit quantization. We also give an initial look into how high-norm tokens holding global information behave under quantization and how that affects anomaly detection. Future works include further investigation into how the representations learned by high-norm tokens and other patch tokens behave at lower precision, and into the value of data augmentation and large-scale pretraining for quantization in OOD.

{
    \small
    \bibliographystyle{ieeenat_fullname}
    \bibliography{robustness,architectures,quantization,datasets,attention,concepts}
}

\end{document}